\documentclass[]{interact}

\usepackage[singlelinecheck=off]{caption}
\usepackage[utf8]{inputenc}
\usepackage{svg}
\usepackage{graphics}
\usepackage{multirow}
\usepackage[english]{babel}

\usepackage{algorithm}
\usepackage{algorithmic}

\usepackage{geometry}
\usepackage{pdflscape}

\usepackage{epstopdf}
\usepackage{subfigure}
\usepackage{hyperref} 
\hypersetup{hypertex=true,
            colorlinks=true,
            linkcolor=blue,
            anchorcolor=blue,
            citecolor=blue}

\usepackage[super,square,numbers,sort&compress]{natbib}
\bibpunct[, ]{(}{)}{;}{a}{}{,}
\renewcommand\bibfont{\fontsize{10}{12}\selectfont}

\theoremstyle{rusnat}

\theoremstyle{definition}

\theoremstyle{remark}

\begin{document}

\captionsetup[table]{name={Table},labelsep=space}
\captionsetup[figure]{name={Figure},labelsep=space} 

\title{Hybrid intelligence for dynamic job-shop scheduling with deep reinforcement learning and attention mechanism}

\author{
\name{Yunhui Zeng\textsuperscript{a}, Zijun Liao\textsuperscript{b}, Yuanzhi Dai \textsuperscript{b}, Rong Wang\textsuperscript{b}, Xiu Li\textsuperscript{a} and Bo Yuan\textsuperscript{a}\thanks{Contact Author: Bo Yuan. Email: boyuan@ieee.org}}
\affil{\textsuperscript{a}Shenzhen  International  Graduate  School,  Tsinghua  University, Shenzhen 518055, China; \textsuperscript{b}School  of  Intelligent  Systems  Science  and  Engineering, Jinan University, Zhuhai 519070, China}
}

\maketitle

\begin{abstract}
The dynamic job-shop scheduling problem (DJSP) is a class of scheduling tasks that specifically consider the inherent uncertainties such as changing order requirements and possible machine breakdown in realistic smart manufacturing settings. Since traditional methods cannot dynamically generate effective scheduling strategies in face of the disturbance of environments, we formulate the DJSP as a Markov decision process (MDP) to be tackled by reinforcement learning (RL). For this purpose, we propose a flexible hybrid framework that takes disjunctive graphs as states and a set of general dispatching rules as the action space with minimum prior domain knowledge. The attention mechanism is used as the graph representation learning (GRL) module for the feature extraction of states, and the double dueling deep Q-network with prioritized replay and noisy networks (D3QPN) is employed to map each state to the most appropriate dispatching rule. Furthermore, we present Gymjsp, a public benchmark based on the well-known OR-Library, to provide a standardized off-the-shelf facility for RL and DJSP research communities. Comprehensive experiments on various DJSP instances confirm that our proposed framework is superior to baseline algorithms with smaller makespan across all instances and provide empirical justification for the validity of the various components in the hybrid framework.

\end{abstract}


\section{Introduction}
\label{1}
The job-shop scheduling problem (JSP) aims to determine the optimal sequential assignments of machines to jobs consisting of a series of operations subject to one or more criteria such as production cost and makespan. It is a typical combinatorial optimization problem closely related to industrial scheduling and production control (Satyro et al. \citeyear{z1}) and has been proven to be non-deterministic polynomial-time hard (NP-hard) (Garey, Johnson, and Sethi \citeyear{z2}) when the number of machines is greater than two. Furthermore, the emergence of smart manufacturing calls for effective solutions for the dynamic job-shop scheduling problem (DJSP) to confront the variabilities in production requirements, consumption of time, and available machines in the environment (Mohan, Lanka, and Rao \citeyear{mohan2019review}).

Among the existing solutions of DJSP, dispatching rules (Haupt \citeyear{z4}) such as Most Operation Remaining (MOR) and Shortest Processing Time (SPT) are designed based on practical experiences and can react to dynamic events promptly. However, there is a lack of assurance on their performance, and each of them is only suitable for specific scenarios. For example, MOR achieves lower makespan than SPT on ft06, an instance in the OR-Library (Beasley \citeyear{z5}), but SPT outperforms MOR on another instance la01. Consequently, the human operator needs to consistently choose one of the multiple commonly-used dispatching rules based on their own knowledge, which is impractical and tedious in most scenarios. Mathematical programming (Manne \citeyear{z6}) can find the optimal solution by solving an optimization problem with constraints. However, it suffers from the curse of dimensionality, and it is intractable to find the optimal scheduling in real-time, even for small-size instances. By contrast, meta-heuristic algorithms such as genetic algorithms (Gonçalves, Magalhães Mendes, and Resende \citeyear{z7}) and simulated annealing (Van  Laarhoven, Aarts, and Lenstra \citeyear{z8}) aim to find a locally effective strategy by decomposing the DJSP into a series of static sub-problems and solving them separately (Luo \citeyear{z9}). The major issue is that these techniques feature poor generalization ability and need to be redesigned for even minor changes to the original problem. 

With the development of artificial intelligence, data-driven approaches such as deep learning (Deng and Yu \citeyear{z10}) and machine learning (Das et al. \citeyear{z11}) have attracted great interest from researchers. Most of these methods rely on fully labeled datasets to train models and deploy them to unseen scheduling scenarios. With sufficient computing power and high-quality data sources, various encouraging results have been achieved following this supervised manner (Tian and Zhang \citeyear{z12}; Teymourifar et al. \citeyear{z13}). Unfortunately, for real-world production systems, it is often infeasible to obtain large-scale and high-quality training data.

Reinforcement learning (RL) aims at designing optimal policies to maximize the cumulative rewards over time through trial-and-error interactions with environments. It has achieved great success in many domains, such as GO (Silver et al. \citeyear{z14,z15}), self-driving cars (Kendall et al. \citeyear{z16}), biology (Senior et al. \citeyear{z17}), surgical robotics (Richter, Orosco, and Yip \citeyear{z18}) and smart manufacturing (Xia et al. \citeyear{z19}). For the DJSP, Wang and Usher (\citeyear{z20}) applied Q-learning to select self-designed dispatching rules on a single machine. The combination of deep learning and RL creates the field of deep reinforcement learning (DRL), which has been used to solve the JSP in recent years. Turgut and Bozdag (\citeyear{z21}) applied the deep Q-network (DQN) to schedule jobs dynamically to minimize the delay time of jobs based on a discrete event simulation experiment. Wang et al. (\citeyear{z22}) developed an improved Q-learning called dual Q-learning (D-Q)method for the assembly job shop scheduling problem (AJSSP). The top level Q-learning tries to learn the dispatching policy that can minimize machine idle rate, and the bottom level Q-learning tries to find the optimal scheduling policy to minimize the overall earliness of all jobs. Wang et al. (\citeyear{z3}) employed proximal policy optimization (PPO) to alleviate the dimension disaster caused by the increase of problem size and obtain real-time production scheduling. Hameed and Schwung (\citeyear{z23}) described a distributed RL method and applied deep deterministic policy gradient(DDPG) to each agent so that each agent optimizes its own cumulative reward. 

A critical step in applying RL to the JSP is to formulate it as a Markov decision process (MDP), involving the key definitions of action, state and reward. However, previous studies often introduce customized formulations based on their own domain knowledge of JSP with handcrafted engineering and workload-specific implementation, making their experimental results incomparable. For the action space, some methods select an action from a fixed operation set without any constraints (Turgut and Bozdag \citeyear{z21}; Luo et al. \citeyear{z24}; Qu, Wang and Shivani \citeyear{z25}), while others may select an action from the eligible operation set to ensure that the action can be directly executed (Zhang et al. \citeyear{z30}). There are also cases where self-designed (Wang et al. \citeyear{z22}; Wei and Zhao \citeyear{z26}) or customized dispatching rules (Lin et al. \citeyear{z27}; Yang and Yan \citeyear{28}; Liu, Chang, and Tseng \citeyear{z28}) are designated as the actions. For the state space, some studies define the state as a matrix composed of customer order features and system features, such as the number of machines and the average completion time (Liu, Chang, and Tseng \citeyear{z28}), without specifying the relationship among different entities. Other studies use the disjunctive graph to represent the original state and the graph neural network (GNN) to do the feature extraction (Zhang et al. \citeyear{z30}). A major benefit is that they are capable of dealing with instances of varying sizes without extra training. However, the performance of the GNN may collapse and the computational cost may increase dramatically for complex problems (Wu et al.\citeyear{z31}), as the increasing number of neighborhood nodes may propagate noisy information (Zhou et al. \citeyear{z32}). 

In summary, existing approaches deeply rely on the knowledge and experience of domain experts, which can be inevitably subjective and require significant implementation efforts. Meanwhile, there is a lack of common benchmarks and some studies even did not conduct experimental comparisons with other RL methods. Furthermore, the performance of RL-based scheduling systems is sensitive to the quality of formulation. Consequently, we argue that there is a critical need for a general, versatile and effective formulation scheme as well as a high-quality benchmark that is friendly to researchers and practitioners. Table \ref{Table.1} presents the details of existing RL-based JSP methods where FJSP, DFJSP, and FSP are the flexible job-shop scheduling problem, the dynamic, flexible job-shop scheduling problem, and the flow-shop scheduling problem, respectively. 

\begin{table}[h]\Huge
\caption{Existing RL-based methods for job shop scheduling.}
\label{Table.1}
\resizebox{\textwidth}{!}{
\begin{tabular}{cccccccc}
\hline
\textbf{Work}& \textbf{\begin{tabular}[c]{@{}c@{}}State \\ representation\end{tabular}} & \textbf{\begin{tabular}[c]{@{}c@{}}Action \\ space\end{tabular}} & \textbf{RL algorithm}& \textbf{Objective}& \textbf{Dynamic events}& \textbf{Problems} & \textbf{\begin{tabular}[c]{@{}c@{}}Comparison with \\ other RL methods\end{tabular}} \\ \hline
Park et al. \citeyearpar{z29}& GNN& Eligible operations& PPO& Makespan& None& JSP& Yes\\Hameed and Schwung \citeyearpar{z23}& GNN& Operations& PPO,DDPG& Makespan& None& JSP& No\\Zhang et al. \citeyearpar{z30}& GNN& Eligible operations& PPO& Makespan& None& JSP& No\\Lin et al. \citeyearpar{z27}& Matrix& Customized dispatching rules&multiclass DQN& Makespan& None& JSP& No\\Wang et al. \citeyearpar{z22}& Matrix& Self-designed rules& dual Q-learning& \begin{tabular}[c]{@{}c@{}}Total weighted earliness penalty, \\ makespan \end{tabular} & None& AJSSP& Yes\\Wang et al. \citeyearpar{z3}& Matrix& Eligible operations& PPO& Makepsan& \begin{tabular}[c]{@{}c@{}}Machine breakdown,\\Job rework\end{tabular} & DJSP& No\\Luo et al.  \citeyearpar{z24}& Matrix& Operations& DQN & Makespan& New job insertions& DJSP& Yes\\ Luo \citeyearpar{z9}& Matrix& Self-designed rules& DQN& Tardiness& New job insertions& DFJSP& Yes\\
Turgut and Bozdag \citeyearpar{z21}& Matrix& Operations& DQN& Job Delay time& New job insertions& DJSP& No\\ Yang and Yan \citeyearpar{28}& Matrix& Customized dispatching rules& B-Q learning& Tardiness& New job insertions& DJSP& No\\
Wei and Zhao\citeyearpar{z26}& Matrix& Self-designed rules& Q-learning& Tardiness& Job delay& DJSP& No\\ Qu, Wang, and Shivani \citeyearpar{z25}  & Matrix& Operations& \begin{tabular}[c]{@{}c@{}}DQN\end{tabular} & \begin{tabular}[c]{@{}c@{}}On-time delivery,\\the wait-in-process products\end{tabular}& New job insertions& DJSP& No\\ Arviv, Stern, and Edan \citeyearpar{ta1} & Matrix& Operations& dual Q-learning& Makespan& None& FSP& No\\
Reyna and Mart{\'\i}nez-Jim{\'e}nez, \citeyearpar{ta2}& Matrix& Operations& Q-learning& Makespan& None& FSP& No\\ Liu, Chang, and Tseng \citeyearpar{z28}&Matrix& Customized dispatching rules& DDPG & Makepsan& New job insertions& DJSP& Yes\\ Ours & \textbf{Attention mechanism}& \textbf{General dispatching rules}& \textbf{D3QPN}& Makespan& \textbf{\begin{tabular}[c]{@{}c@{}} Machine breakdown,\\ different order requirements\end{tabular}}& DJSP& \textbf{Yes}\\ \hline
\end{tabular}
}
\end{table}

In this paper, we propose a general framework to solve the DJSP based on the RL paradigm without handcrafted engineering. The output of RL for each state is a dispatching rule rather than detailed scheduling operations due to its clarity, ease of implementation, and good interpretability that are essential for the production field. In this setting, our framework can be regarded as a prototype towards hybrid intelligence, combining the formalized human knowledge in the form of structured decision rules and the data driven black-box intelligence exercised by the RL agent for selecting appropriate rules for the current state. The states of DJSP are represented by graphs and the attention mechanism is used as the graph representation learning (GRL) module, which is responsible for effective feature extraction. The extracted features are then used as the input to the double dueling deep Q-network with prioritized replay and noisy networks (D3QPN), which is a value-based RL method with various useful extensions to DQN. To provide a principled DSJP environment for RL-based research, we develop a public benchmark named Gymjsp following the same convention as the OpenAI Gym environment that can significantly reduce the burden of other researchers. The overall procedure of the proposed framework is shown in Figure \ref{Figure 1} and the major contributions of our work are summarised as follows:
\begin{figure}[htp]
    \centering
    \includegraphics[width=13cm]{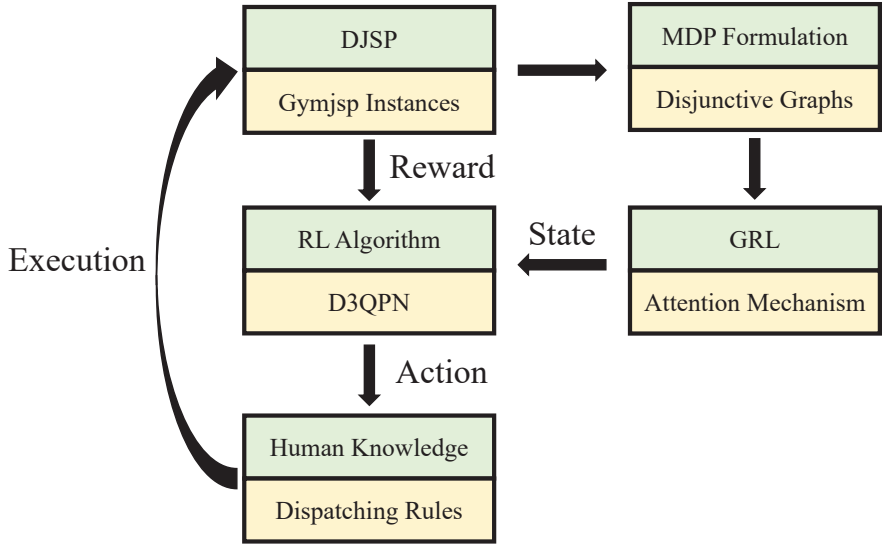}
    \caption{An overview of the proposed framework.}
    \label{Figure 1}
\end{figure}

\begin{itemize}
\item [1.] We effectively formulate the scheduling process of DJSP as an MDP, taking disjunctive graphs as states and general dispatching rules as the action. The disjunctive graph captures the local, global, and even dynamic information of JSP and the general dispatching rules can make our method applicable in a wide range of real smart manufacturing systems. Our formulation method provides a prototype towards hybrid intelligence, which can be an effective paradigm of human-machine cooperation.
\item [2.] We present a benchmark named Gymjsp following the convention of the OpenAI Gym environment to facilitate RL-based DJSP research. To the best of our knowledge, it is the first benchmark specifically targeted at RL-based DJSP.
\item [3.] We use the attention mechanism as the GRL module to extract features from disjunctive graphs. By visualizing the learning process of the attention mechanism, we confirm that the attention mechanism can effectively learn complex relationships among operations in the DJSP.
\item [4.] We propose an improved RL algorithm named D3QPN that has been proved to outperform other RL algorithms in the DJSP with several effective and complementary ingredients.
\end{itemize}

The rest of this paper is organized as follows: Section \ref{2} provides a review of the relevant background, and the problem formulation is specified in Section \ref{3}. The details of our method are introduced in Section \ref{4}, and the experimental results and analyses are presented in Section \ref{5}. This paper is concluded in Section \ref{6} with some discussions and directions for future work.\par

\section{Background}
\label{2}
\subsection{Reinforcement learning}
\label{2.1}
RL is one of the major areas of machine learning, along with supervised learning and unsupervised learning, which is inspired by biology and neuroscience. It is concerned about sequential decision making over time through trial-and-error interactions with the environment and the received reinforcement signals. In general, RL tasks are formulated as a MDP described by a tuple $(\boldsymbol{S}, \boldsymbol{A}, \boldsymbol{P}, \boldsymbol{R})$. At each decision point $t$, the agent will observe the state $s_{t} \in \boldsymbol{S}$ and take a action $a_{t} \in \boldsymbol{A}$ according to the policy $\pi(\boldsymbol{S} \rightarrow \boldsymbol{A})$, after which it enters a new state $s_{t+1}$ with transition probability $p\left(s_{t+1} \mid s_{t}, a_{t}\right) \in \boldsymbol{P}(\boldsymbol{S} \times \boldsymbol{A} \rightarrow \boldsymbol{S})$. Meanwhile, an immediate reward signal $r_{t} \in \boldsymbol{R} (\boldsymbol{S} \times \boldsymbol{A} \times \boldsymbol{S} \rightarrow \mathbb{R})$ is obtained as a result of the state transition $\left(s_{t}, a_{t}, s_{t+1}\right) $ (Sutton and Barto \citeyear{sutton2018reinforcement}). The goal of RL is to find an optimal policy $\pi^{*}$ that maximizes the cumulative rewards:
\begin{equation}
\pi^{*}=\underset{\pi}{\operatorname{argmax}} \mathbb{E}[\sum_{t=0}^{+\infty} r_{t+1} \mid \pi]
\end{equation}
There are two main learning paradigms in RL: value-based methods and policy-based methods. A value-based method outputs the action-value of the pair $(s,a)$: $Q^{\pi}(s, a)=\mathbb{E}^{\pi}\left[\sum_{t=0}^{+\infty} \gamma^{t} r\left(s_{t}, a_{t}\right)\right]$, where $\gamma$ is the discount factor. By contrast, the policy-based method directly searches for an optimal policy using the gradient-based method and outputs the probability of each action. There is also a hybrid scheme called actor-critic where the actor interacts with the environment and updates the policy parameters, and the critic evaluates the policy's value to update the actor parameters, aiming to eliminate all the drawbacks from both value-based and policy-based methods.

\subsection{Graph representation learning}
\label{2.2}
A graph is a structure of objects (nodes or vertices), along with a set of relationships (edges or links) between each pair of objects. GRL aims to map a graph to a finite-dimensional vector space that captures the key features of the graph and assists with subsequent processing and decision making. One of the simplest methods to extract the key information from the graph is combining the features of nodes directly into a matrix, which can be viewed as a summarized graph feature. However, it cannot capture the relationship between each pair of nodes and requires manual feature filtering based on specific domain knowledge. Otherwise, the redundant information may be harmful to subsequent processing and decision making. 

The GNN is a deep learning-based GRL model (Zhou et al. \citeyear{z32}) that has been widely applied in GRL recently (Cui et al. \citeyear{n2}; Chen et al. \citeyear{n3}), which can help nodes incorporate high-order neighborhood relationships \citep{shi2020evolutionary}. There is a hidden embedding $\mathbf{h}_{u}^{(k)}$ corresponding to each node $u$ in a GNN, which is updated  to generate $\mathbf{h}_{u}^{(k+1)}$ by aggregating the message $\mathbf{m}_{\mathcal{N}(u)}$ from the graph neighborhood nodes $\mathcal{N}(u)$ iteratively. The update function can be expressed as:
\begin{equation}
\begin{aligned}
\mathbf{h}_{u}^{(k+1)} &=\operatorname{F}^{(k)}\left(\mathbf{h}_{u}^{(k)}, \text {G}^{(k)}\left(\left\{\mathbf{h}_{v}^{(k)}, \forall v \in \mathcal{N}(u)\right\}\right)\right) \\
&=\operatorname{F}^{(k)}\left(\mathbf{h}_{u}^{(k)}, \mathbf{m}_{\mathcal{N}(u)}^{(k)}\right)
\end{aligned}
\end{equation}
where both $F$ and $G$ are differentiable functions, while $F$ is to update $\mathbf{h}_{u}^{(k)}$ and $G$ is to aggregate message $\mathbf{m}_{\mathcal{N}(u)}$. After $K$ iterations, the embeddings of each node can be defined using the outputs of the final layer $\mathbf{z}_{u}=\mathbf{h}_{u}^{(K)}$, $\forall u \in \mathcal{V}$ (Hamilton \citeyear{n1}). Note that, the computation on $\mathcal{N}(u)$ can lead to dramatically increased time consumption given a complex graph, resulting in deteriorated performance.

The attention mechanism can also be extended to GRL. Its main idea is to aggregate a variable-sized input by a function while focusing on the most relevant parts of the input to make decisions. Each node in the set of neighborhood nodes $N(t)$ of the target node $t$ is assigned an attention weight $\alpha_{i}$ using the attention-based GRL method, where $\alpha$ is from 0 to 1 with $\sum_{i \in N(t)} \alpha_{i}=1$. The attention mechanism can identify the information in the input graph pertinent to subsequent tasks by calculating the importance of different nodes. Furthermore, many operations in attention can be performed in parallel, greatly improving computational efficiency.

\subsection{Simulation environment for DJSP}
\label{2.3}
The research on JSP has significantly benefited from several simulation environments containing standard instance sets for simulating real-world scheduling environments and evaluating different algorithms. For example, Demirkol's instances (Demirkol, Mehta, and Uzsoy \citeyear{z38}) are for the problems of maximum lateness and minimum makespan in job shops and flow shops. Taillard \citeyearpar{z39} created instances for the job shop, open shop, and permutation flow shop scheduling problems with the objective of minimizing the makespan. In addition, the well-known instance set, OR-Library, has been widely adopted as the test environment of JSP in recent years. These instances contain the key parameters and data of JSP, such as the processing time of each operation, the number of jobs, the number of machines, etc. However, these simulation environments were originally designed for traditional methods such as dispatching rules, which cannot be directly used for RL-based approaches. Consequently, researchers need to adapt these environments accordingly to suit RL methods, which is challenging and time-consuming. Furthermore, it is common that researchers tend to repackage the environments based on their own domain knowledge and specific purposes. For example, the information used to define the state is usually different: some researchers take into account the consumption of time and status of operations (Zhao and Zhang \citeyear{z40}), while others may consider the status of jobs and machines when designing their states (Wang et al. \citeyear{z3}). For the DJSP, researchers may set different effects (e.g., the length of time delays) for the same dynamic event (e.g., machine breakdown), which makes the experimental results incomparable. In summary, there is a lack of a public, easy-to-use, and standardized benchmark that can help promote the development of RL in the DJSP. 

\section{Problem formulation}
\label{3}
\subsection{DJSP formulation}
\label{3.1}
The DJSP is a dynamic allocation problem in a resource-limited environment facing variability in production requirements, consumption of time, available machines and unexpected faults. In each DJSP, there are $m$ machines $M = \{M_1, M_2 ... M_m \}$ and $n$ jobs $J = \{J_1, J_2 ... J_n \}$. Each job has $m$ operations to be processed where the operation sequence of $J_i$ is defined as $O_i = \{O_{i1}, O_{i2}..O_{il}... O_{im}\}$. The machine matrix $MO =\left\{M_{i l} \mid M_{i l}= 0,M_{1}, M_{2}, \cdots, M_{m}\right\}(i=1,2, \ldots, n, l=1,2, \ldots, m)$ indicates that $O_{il}$ should be processed by $M_{il}$. The value of $T_{il}$ in the processing time matrix $TO=\left\{T_{i l} \mid T_{i l} \ge 0\right\}(i=1,2, \ldots, n, l=1,2, \ldots, m)$ specifies the required processing time of $O_{il}$. If $M_{i l} = 0$, then $T_{i l} = 0$, which indicates that job $J_{i}$ does not require the operation $O_{i l}$ (Wang et al. \citeyear{z3}).

In our work, the goal of DJSP is to minimize the makespan with the following objective function:

\begin{equation}
\min \left\{\max _{1 \leq i \leq n} C_{i l}\right\}   
\end{equation}
where $C_{il}$ is the completion time of operation $O_{il}$ on machine $m_{il}$. As the machine matrix $MO$ and the processing time matrix $TO$ are given in advance, the completion time $C_{i l_{i l}}$ can be given as:

\begin{equation}
\begin{aligned}
&C_{i l}=\max \left(C_{i(l-1)}, C_{a b}\right)+T_{wait}+T_{i l} \\&
M_{i l}=M_{a b}; i,a=1, \ldots,n; b=1, \ldots,m; l=2, \ldots, m
\end{aligned}
\end{equation}
where $C_{i(l-1)}$ is the completion time of the previous
operation $O_{i(l-1)}$ in the same job $J_{i}$; $C_{a b}$ is the completion time of the previous operation $O_{a b}$ on the same machine $M_{i l}$; $T_{wait}$ indicates the time that it takes to keep machine $M_{i l}$ waiting without loading operation $O_{i l}$ when the machine $M_{i l}$ is idle, which is ineffective in most cases.

\begin{figure}[htp]
    \centering
    \includegraphics[width=12cm]{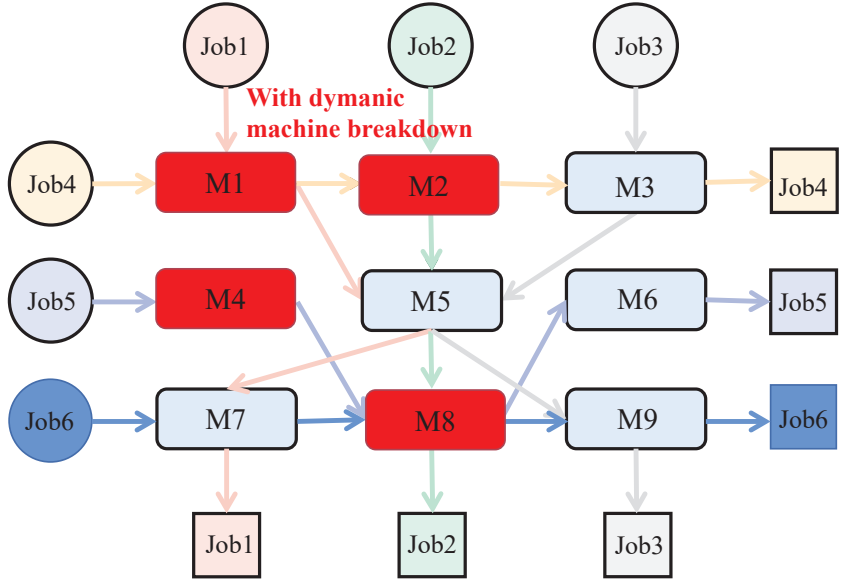}
    \caption{A simple example of DJSP with nine machines and six jobs where the arrows of the same color indicate the unique flow of a specific job.}
    \label{Figure 2}
\end{figure}

The entire process of a simple DJSP example is shown in Figure \ref{Figure 2}, where machines in red have the possibility of breakdown. From the circular starting point to the square ending point, each job is processed by a certain sequence of machines, resulting in a unique route of completion. The simultaneous execution of multiple jobs may inevitably lead to competition for machine resources, blocking the execution of jobs and increasing the total time cost.

The assumptions in the DJSP are listed as follow:
\begin{itemize}
\item [1.] Each machine can conduct only one operation at a time.
\item [2.] Each operation of a job can be executed by only one machine at a time.
\item [3.] Each operation of a job cannot be executed until its primary operations are completed.
\item [4.] Each operation that has been started cannot be suspended or terminated.
\item [5.] All the operations of the same job need to follow a unique sequence.
\item [6.] All the transportation times and setup times are negligible.
\item [7.] The machine matrix $MO$ and processing time matrix $TO$ are known in advance.
\item [8.] There is some possible disturbance in the production environment, such as machine breakdown.
\end{itemize}

\subsection{MDP formulation for DJSP}
\label{3.2}
The DJSP aims to determine the order of operations in a dynamic environment, which is essentially a sequential decision problem and can be formulated as an MDP defined by states, actions, and reward functions. 

States describe a common feature set that can represent the characteristics related to the environment and the objective of scheduling. We use the disjunctive graph whose details can be found in Section \ref{3.3} as the abstract of the local, global, and even dynamic information of the scheduling environment. Furthermore, a disjunctive graph with a numerical representation of the state attributes is capable of specifying all states of different scheduling problems. In practice, we can rely on domain knowledge to extract features from disjunctive graphs manually or use GRL methods such as the GNN for automatic feature extraction.

\begin{figure}[h!]
    \centering
    \includegraphics[width=15cm]{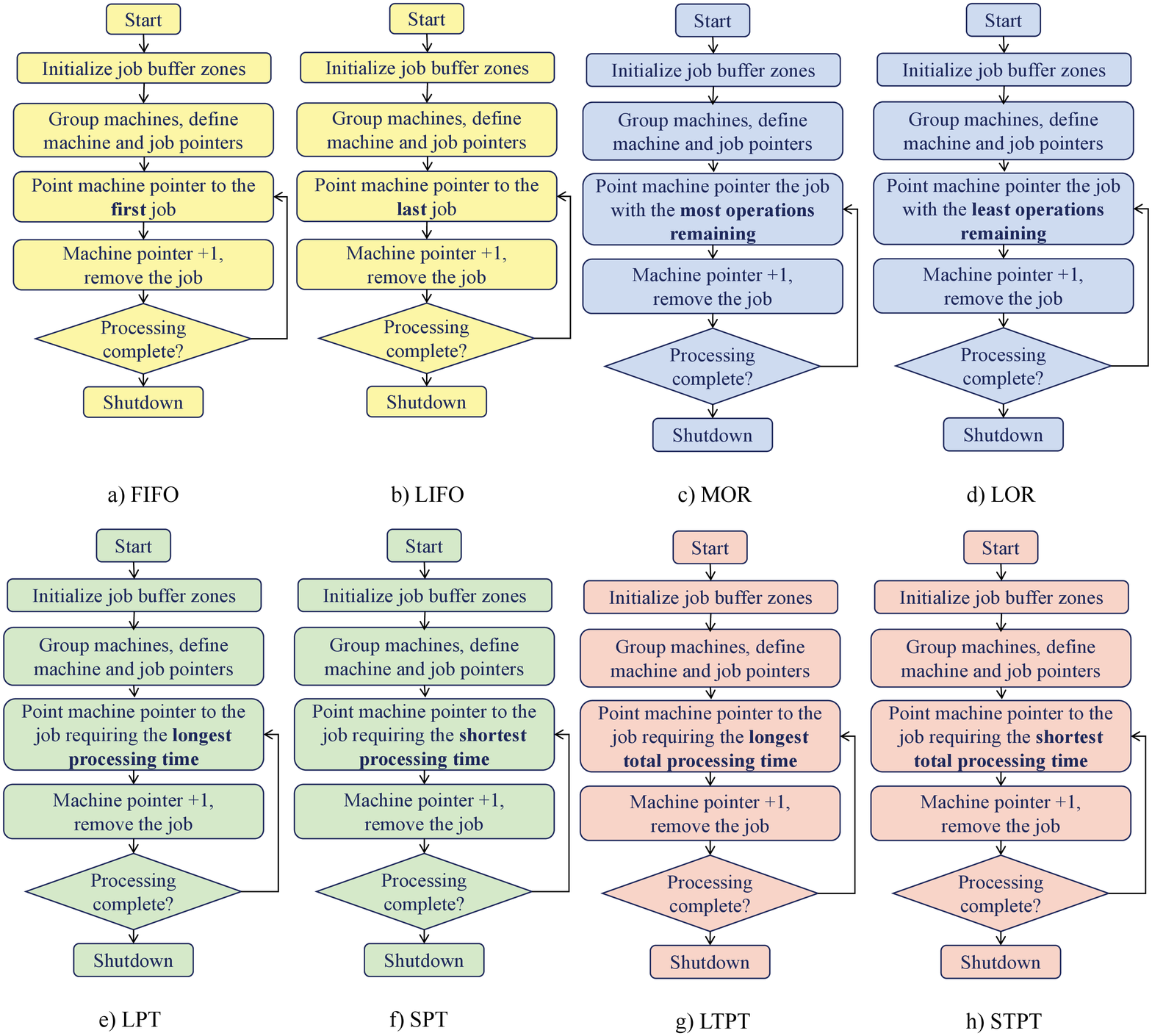}
    \caption{The flowcharts of eight rule-based scheduling algorithms: (a) First In First Out (FIFO); (b) Last In Last Out (LIFO); (c) Most Operations Remaining (MOR); (d) Least Operations Remaining (LOR); (e) Longest Processing Time (LPT); (f) Shortest Processing Time (SPT); (g) Longest Total Processing Time (LTPT); (h) Shortest Total Processing Time (STPT).}.
    \label{Figure 3}
    
\end{figure}

Traditionally, the action at each step is a detailed operation to be directly executed in the environment, which can be unsafe for RL-based methods due to the black-box nature of neural networks. Instead, we adopt dispatching rules as the action space of our RL method for the sake of simplicity, ease of implementation, and good interpretability, which are desirable for management purposes. As an analogy to the factory staff who routinely choose one of the dispatching rules based on their domain knowledge, our RL-based method strategically selects the most appropriate dispatching rule according to the current state. In this way, it is possible to overcome the limitation of any single dispatching rule and can be seen as a hybrid intelligence method. Figure \ref{Figure 3} shows the execution flows of eight dispatching rules adopted in our work. Note that, in addition to rule-based systems, we can use any scheduling methods as possible actions such as meta-heuristic and RL-based algorithms.

The definition of reward should be closely associated with the scheduling objective. Although the goal of the DJSP is to minimize the makespan, it can only be obtained when the entire scheduling process is over, resulting in the challenging sparse reward issue (Rauber et al. \citeyear{rauber2021reinforcement}). Since the makespan is closely related to machine utilization (in general, the higher the machine utilization, the smaller the makespan), our reward function is defined as:
\begin{equation}
reward = -\frac{\sum^k_1 \frac{\text{number of idle machines}} {\text{number of total machines}}} {k}
\end{equation}
where $k$ is a hyperparameter controlling the number of steps for which the output dispatch rule is used. Although it is possible to design the reward function in more sophisticated ways, our purpose is to show that, with such a simple reward function, the proposed hybrid RL framework can already achieve superior performance, with plenty of room for further improvement.

\subsection{State representation}
\label{3.3}
\begin{figure}[h!]
    \centering
    \includegraphics[width=15cm]{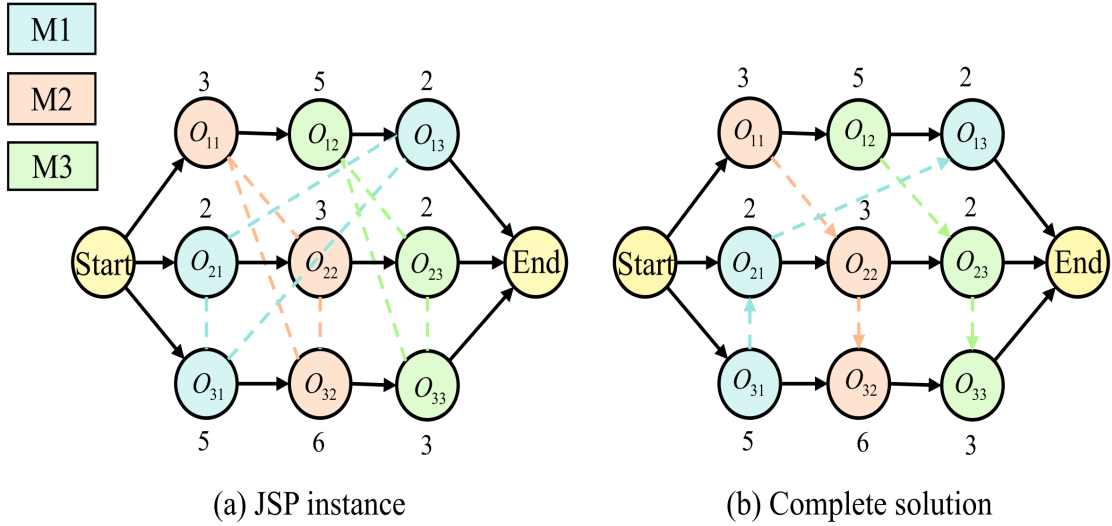}
    \caption{An example of a 3×3 JSP instance. (a) Initial state. Each circle is an operation and $O_{ij}$ is the $j^{th}$ operation of the $i^{th}$ job. The number on each circle represents the consumption of time of the operation. The solid and dotted lines are conjunctive and disjunctive edges, respectively. Vertices of the same color indicate that the operations are processed on the same machine; (b) Its complete solution.}
    \label{Figure 4}
\end{figure}

The performance of RL on the JSP relies heavily on the quality of hand-crafted feature representation. The disjunctive graph $G=(V, C \cup D)$ is a popular representation for JSP instances (Balas \citeyear{z41}). $V$ is the set of all possible operations represented as vertices in the graph, including two dummy vertices: the start and terminal ones, both of which have zero consumption of time. $C$ is a set of directed edges (conjunctions) representing the precedence constraint between two consecutive operations in the same job. $D$ is the undirected edge set (disjunctions) representing the machine-sharing constraint between two vertices. When a pair of operations can be processed on the same machine, their corresponding vertices are connected with an undirected edge. Therefore, solving a JSP instance is equivalent to turning every undirected disjunctive edge into a directed edge so that the result is a directed acyclic graph. Figure \ref{Figure 4} gives an example of a simple JSP instance and its solution.

Although the disjunctive graph contains the static information in the JSP, such as precedence constraints and machine-sharing constraints, it fails to represent the dynamic state information in the DJSP. To address this issue, we add several features into the vertices in $G$ as follows:
\begin{itemize}
\item [1.] Job ID: The sequence numbers of jobs that contain this operation.
\item [2.] Step ID (Operation ID): The sequence number of this operation.
\item [3.] Node type: -1 for unfinished, 0 for in the process, and 1 for completed.
\item [4.] Completion ratio: The completion rate of the whole job when this operation is completed.
\item [5.] Consumption of time: The required time to complete the operation.
\item [6.] Number of remaining operations: The number of subsequent operations.
\item [7.] Waiting time: The time elapsed from the beginning until this operation can be processed.
\item [8.] Remaining time: The remaining consumption of time (0 for not in processing).
\item [9.] Doable: $True$ if this operation is doable.
\item [10.] Machine ID: The sequence number of the machine that can process this operation (0 if this operation cannot be processed).
\end{itemize}

\section{Methodology}
\label{4}
\subsection{Proposed architecture for DJSP}
\label{4.1}
This paper proposes a hybrid framework to solve the DJSP in smart manufacturing using the attention mechanism and RL, as shown in Algorithm \ref{algorithm 1}. The attention mechanism is a GRL module to transfer the original state $s$, which is a disjunctive graph, to the extracted state $s_{a}$, which is a vector, and D3QPN is used to learn the action value $Q(s_{a}, \cdot,\theta)$. In Algorithm \ref{algorithm 1}, given a DJSP, the system initially formulate the DJSP as an MDP $(\mathbf{S}, \mathbf{A}, \mathbf{P}, \mathbf{R})$, where $\mathbf{S}$ is the disjunctive graph (Line 2). Then, the attention mechanism transfers this disjunctive graph to a vector that can be directly input to D3QPN (Line 4). With the sampled noisy network $\xi$ (Line 5) available, we retrieve the dispatching rule corresponding to the maximal value (Line 6). This dispatching rule is executed for $k$ steps to make the control process more stable (Line 7 to Line 9) and the transition $\left(\mathbf{s}_{\mathbf{a}}, \mathbf{a}, \mathbf{r}, \mathbf{s}_{\mathbf{a}}^{\prime}\right)$ is stored in the replay buffer with maximal priority to make sure that each experience is seen at least once. The RL algorithm and GRL parameters are updated using the mini-batch sampled from the replay buffer. The above procedure is repeated until the maximal number of epochs $max\_epoch$ is reached. 

\begin{algorithm}
	\renewcommand{\algorithmicrequire}{\textbf{Input:}}
	\renewcommand{\algorithmicensure}{\textbf{Output:}}
	\caption{Proposed framework using D3QPN and attention mechanism}
	\label{algorithm 1}
	\begin{algorithmic}[1]
	\REQUIRE Environment $Env$, set of random variables of the network $\epsilon$
	\STATE Initialize the noisy behavior network $\mathrm{Q}$ with random weights $\theta$; GRL module; replay buffer; exponent $\alpha$ of prioritized replay; minibatch $\tau$.
	\STATE Formulate the DJSP as an MDP$(\mathbf{S}, \mathbf{A}, \mathbf{P}, \mathbf{R})$, where $\mathbf{S}$ is a disjunctive graph.
	\FOR{epoch number $epoch=1,2, \ldots max\_epoch$}{
	    \STATE Extract the state $s_a$ from $s$ using the GRL module (Algorithm \ref{algorithm 2}).
	    \STATE Sample a noisy network $\xi \sim \varepsilon$.
	    \STATE Select the dispatching rule $a \leftarrow \operatorname{argmax}_{a_{t} \in A} Q(s_{a}, \cdot,\xi, \theta)$.
	    \FOR{schedule cycle $i=1,2, \ldots k$}
	    {   \STATE  Execute dispatching rule $a$ and observe new disjunctive graph $s'$.
	        \STATE Extract the state $s'_a$ from $s'$ using Algorithm \ref{algorithm 2}.
	    }\ENDFOR
	    \STATE Store $\left(\mathbf{s}_\mathbf{a}, \mathbf{a}, \mathbf{r}, \mathbf{s}_ \mathbf{a}^{\prime}\right)$ in replay buffer with maximal priority $p_{t}=\max _{i<t} p_{i}$.
	    \STATE Sample the minibatch of transitions with probability $Per(j)=p_{j}^{\alpha} / \sum_{\tau} p_{\tau}^{\alpha}$.
	    \STATE Update the learning algorithm and GRL module based on Algorithm \ref{algorithm 3}.
	}\ENDFOR
	\ENSURE The learned RL module and the GRL module
	\end{algorithmic}  
\end{algorithm}

\subsection{Graph representation learning using attention mechanism}
\label{4.2}
\begin{figure}[htp]
    \centering
    \includegraphics[width=13.5cm]{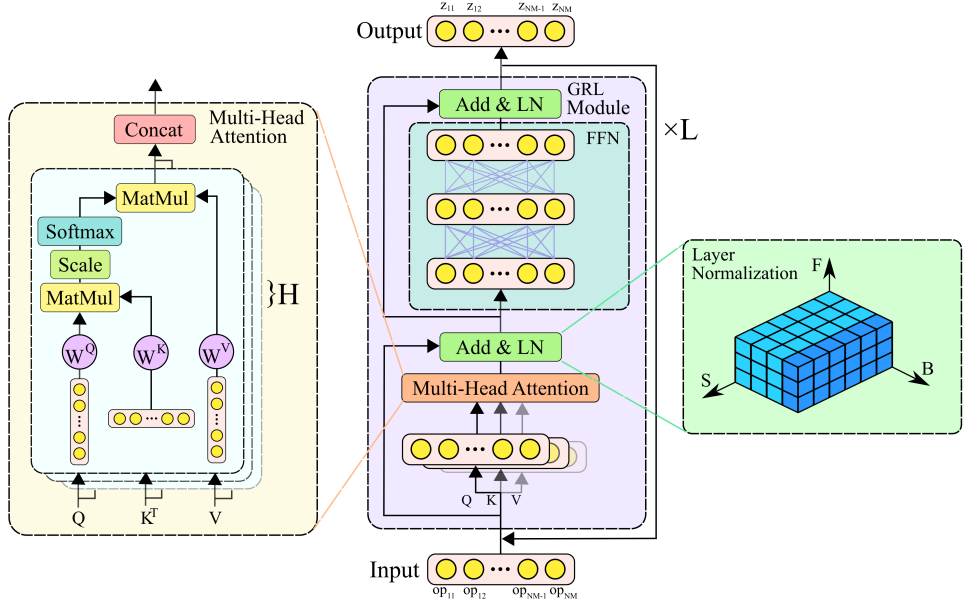}
    \caption{Overall structure of the proposed graph representation learning module.}
    \label{Figure 5}
\end{figure}

Our GRL module uses stacked self-attention, residual connection (He et al. \citeyear{z43}), layer normalization (LN) (Ba, Kiros, and Hinton \citeyear{z44}) and the fully-connected feed-forward network (FFN) to map the sequence that contains all operations $X^0=(x^0_{11}, x^0_{12}, \dots, x^0_{nm-1}, x^0_{nm})$ in the input disjunctive graph to a sequence of extracted features $X^L=(x^L_{11}, x^L_{12}, \dots, x^L_{nm-1}, x^L_{nm})$, which has the same shape as the input sequence, where $L$ is the number of GRL module layers. The overall structure of the proposed GRL module is shown in Figure \ref{Figure 5}.

Self-attention is a function that maps a query and a set of key-value pairs to an output. It learns three linear projections to map the state to $Q, K, V \in \mathbb{R}^{d_{length} \times d_{feature}}$, where $d_{length}$ is the number of total operations, and $d_{feature}$ is the dimension of the features. The query $Q$ indicates the type of operations that the current operation concerns; the key $K$ indicates the type of the current operation; the value $V$ contains the information of this operation. In the first step, we compute the matching scores among all operations by the matrix multiplication of queries and keys. The scores are then divided by $\sqrt{d_{feature}}$ to prevent the attention from being over-focused on nodes with large scores (Vaswani et al. \citeyear{z42}). Next, we apply a softmax function to obtain the attention weight on the values. Finally, we conduct the matrix multiplication of the attention weight and value to obtain the $output \in \mathbb{R}^{d_{length} \times d_{feature}}$. The aforementioned processes can be executed in parallel, which can greatly improve computational efficiency. The procedure of the attention function is as follows:

\begin{equation}
\text{Attention}(Q,K,V)=\text{softmax}(\frac{QK^{\prime}}{\sqrt{d_{feature}}})V
\end{equation}

Multi-head attention acquires more information by integrating the outputs of multiple attention functions with different query matrices and key matrices.

\begin{equation}
\begin{aligned}
\operatorname{MultiHead}(Q, K, V) &=\text {Concat}\left(\text {head}_{1}, \ldots, \operatorname{head}_{\mathrm{H}}\right) W^{multi} \\
\text { where head}_{\mathrm{i}} &=\operatorname{Attention}\left(Q W_{i}^{Q}, K W_{i}^{K}, V W_{i}^{V}\right)
\end{aligned}
\end{equation}

where $W_{i}^{Q},W^K_i,W^V_i \in \mathbb{R}^{d_{feature} \times \frac{d_{feature}}{\text H}}$ and $W^{multi} \in \mathbb{R}^{d_{feature} \times d_{feature}}$ are the parameter matrices, and $H$ is the number of attention heads.

Following the multi-head attention layer, the FFN is composed of two layers of linear transformation, which maps the input from $d_{input}$ dimensions to $d_{output}$ dimensions using the parameter $W_1,W_2,b_1$ and $b_2$. We also apply residual connections around each of the multi-head attention and the FFN to ease the issue of gradient disappearance. After the residual connection, we employ LN, a regularization method, to avoid internal covariate shifts when there are few examples in a single batch. Algorithm \ref{algorithm 2} gives the complete process of the proposed GRL.

\begin{algorithm}
	\renewcommand{\algorithmicrequire}{\textbf{Input:}}
	\renewcommand{\algorithmicensure}{\textbf{Output:}}
	\caption{Graph representation learning using attention mechanism}
	\label{algorithm 2}
	\begin{algorithmic}[1]
	\REQUIRE $X^0=\{x^0_{11},\dots,x^0_{nm}\}$: Disjunctive features
	\STATE Initialize the following parameters:
	\STATE (1)$L$: number of GRL module layers;
	\STATE (2)$H$: number of head of Multi-Head Attention;
	\STATE (3)$W^{Q^{l}},W^{K^{l}},W^{V^{l}}, W^{{multi}^l}, W^l_1, W^l_2$: weight matrix, where $l=1,\dots,L$;
	\STATE (4) $\gamma^l_1, \gamma^l_2, \beta^l_1, \beta^l_2$: parameters for LN function;
	\STATE (5)$b^l_1,b^l_2$: bias for FFN.
	\FOR{$l \leftarrow 1$ to $L$} {
        \FOR{$h \leftarrow 1$ to $H$} {
            \STATE $Q^{l} \leftarrow X^{l-1} \cdot W^{Q^{l}},K^{l} \leftarrow X^{l-1} \cdot W^{K^{l}},V^{l} \leftarrow X^{l-1} \cdot W^{V^{l}}$.
            \STATE $head^{l}_h \leftarrow $softmax$( \frac {Q^{l}\cdot {K^{l}}^{\prime}} {\sqrt{d_{feature}}})V^{l}$.
        }\ENDFOR
        \STATE $A^{l}\leftarrow $Concat $(head^{l}_1, \dots, head^{l}_H) \cdot W^{{multi}^{l}}$.
        \STATE $X^{l} \leftarrow X^{l-1} + A^{l}$.
        \STATE $X^l \leftarrow $ LN$^l_1(X^l)$.
        \STATE $F^l \leftarrow $max$(0, X^l \cdot W^l_1 + b^l_1) \cdot W^l_2 + b^l_2$.
        \STATE $X^{l} \leftarrow X^{l} + F^{l}$.
        \STATE $X^l \leftarrow $ LN$^l_2(X^l)$.
    }\ENDFOR
	\ENSURE $X^L=\{x^L_{11},\dots,x^L_{nm}\}$: Extracted feature
	\end{algorithmic}
\end{algorithm}

\subsection{Double dueling DQN with prioritized replay and noisy networks}
\label{4.3}

The DQN is an important milestone in RL that can achieve competitive performance compared to humans on various tasks. At each step of DQN, the agent selects an action according to the $\epsilon-greedy$ strategy with respect to the action value and adds a transition into the replay buffer. Then, the agent uses a sampled batch from the replay buffer to optimize the behavior neural network to minimize the loss:

\begin{equation}
\left(R_{t+1}+\gamma \max _{a^{\prime}} Q_{\bar{\theta}}\left(s_{t+1}, a_{t+1}\right)-Q_{\theta}\left(s_{t}, a_{t}\right)\right)^{2}
\end{equation}
where $\gamma$ is a discount factor. The behavior network is updated every step, while the target network is updated every specific number of steps to ensure the stability of the action value. In this paper, we propose a double dueling DQN with prioritized replay and noisy networks (D3QPN) algorithm (Algorithm \ref{algorithm 3}), which combines four effective and complementary ingredients with DQN to address the limitations of DQN:

\begin{algorithm}
	\renewcommand{\algorithmicrequire}{\textbf{Input:}}
	\renewcommand{\algorithmicensure}{\textbf{Output:}}
	\caption{The training procedure of D3QPN}
	\label{algorithm 3}
	\begin{algorithmic}[1]
    \REQUIRE step-size $\eta$, target network replacement frequency $N^{-}$, exponent $\beta$ of prioritized replay, replay size $N$
    \STATE Initialize noisy behavior network $\mathrm{Q}$ with random weights $\theta$;
    \STATE Initialize target network $\hat{Q}$ with weights $\theta^{-}=\theta$;
    \STATE Initialize the GRL module $G(\phi)$.
    \FOR{j=1 to $\tau$}{
        \STATE Sample noisy variables $\xi^{\prime} \sim \varepsilon$ for target network .
        \STATE Sample noisy variables $\xi^{\prime \prime} \sim \varepsilon$ for behavior network.
        \STATE Compute importance-sampling weight $w_{j}=(N \cdot Per(j))^{-\beta} / \max _{i} w_{i}$.
        \STATE$\text { Set } y_{j}= \begin{cases}r_{j} & \text { for terminal }  \\ r_{j}+\gamma \max _{a^{\prime}} Q\left(s_{j}, \arg \max _{b \in \mathcal{A}} Q\left(y_{j}, b, \xi^{\prime \prime} ; \theta\right), \xi^{\prime} ; \theta^{\prime}\right) & \text { for non-terminal } \end{cases}$\\
        \STATE \text{ Compute TD-error } $\delta_{j}=\left(y_{j}-Q\left(s_{j}, a_{j} ;\xi; \theta\right)\right)^{2}$. 
        \STATE \text{ Update transition priority } $p_{j}\leftarrow\left|\delta_{j}\right|$.
        \STATE \text { Accumulate weight-change } $\Delta \leftarrow \Delta+w_{j} \cdot \delta_{j} \cdot \nabla_{\theta} Q\left(s_{j}, a_{j}\right)$.
        }
    \ENDFOR
    \STATE Update weights $\phi \leftarrow \phi+\eta \cdot \Delta$, $\theta \leftarrow \theta+\eta \cdot \Delta$, reset $\Delta=0$.
    \STATE Every  $N^{-}$ times, update target network: $\theta^{-} \leftarrow \theta$.
    \ENSURE $Q(\cdot, \varepsilon ; \theta)$ action-value function, GRL module $G(\phi)$
    \end{algorithmic}  
\end{algorithm}

\begin{itemize}
\item [1.] Double Q-learning (Van Hasselt, Guez, and Silver \citeyear{z50}), which is used to reduce harmful overestimations of action value by decoupling the action values used to select and to evaluate an action with the loss:
\begin{equation}
\left(R_{t+1}+\gamma_{t+1} Q_{\bar{\theta}}\left(s_{t+1}, \underset{a^{\prime}}{\operatorname{argmax}} q_{\theta}\left(s_{t+1}, a^{\prime}\right)\right)-Q_{\theta}\left(s_{t}, a_{t}\right)\right)^{2}
\end{equation}

\item [2.] Prioritized replay (Schaul et al. \citeyear{z51}), which is used to sample transitions with probability $p_{t}$ relative to TD error, making the transition that contains more information has a higher probability of being sampled. New transitions are given maximum priority when inserted into the replay buffer to ensure that all transitions are used at least once.  

\item[3.] Dueling networks (Wang et al. \citeyear{z52}), which have two streams to separately estimate the state-value function and the action advantage function for each action and can aggregate the state-action value function without imposing any change to the preceding RL algorithm:
\begin{equation}
q\left(s_{t}, \boldsymbol{a}_{t}\right)=v\left(\boldsymbol{s}_{t}\right)+A\left(\boldsymbol{s}_{t}, \boldsymbol{a}_{t}\right)
\end{equation}

\item[4.] Noisy networks (Fortunato \citeyear{z53}), an effective exploration and exploitation strategy using the factorised Gaussian noise. The parameters of the noisy layer are:
\begin{equation}
\begin{aligned}
&w=\mu^{w}+\sigma^{w} \odot \varepsilon^{w} \\
&b=\mu^{b}+\sigma^{b} \odot \varepsilon^{b}
\end{aligned}
\end{equation}
where the $\mu^{w}, \mu^{b}, \sigma^{w}$ and $\sigma^{b}$ are the parameters of neural network; $\epsilon^{b}$ and $\epsilon^{w}$ are the random variables of factorised Gaussian noise; $\odot$ denotes the element-wise product. The output of the noisy layer is $y=w x+b$, which is the same as the standard fully connected layer.
\end{itemize}

\begin{figure}[htp]
    \centering
    \includegraphics[width=15cm]{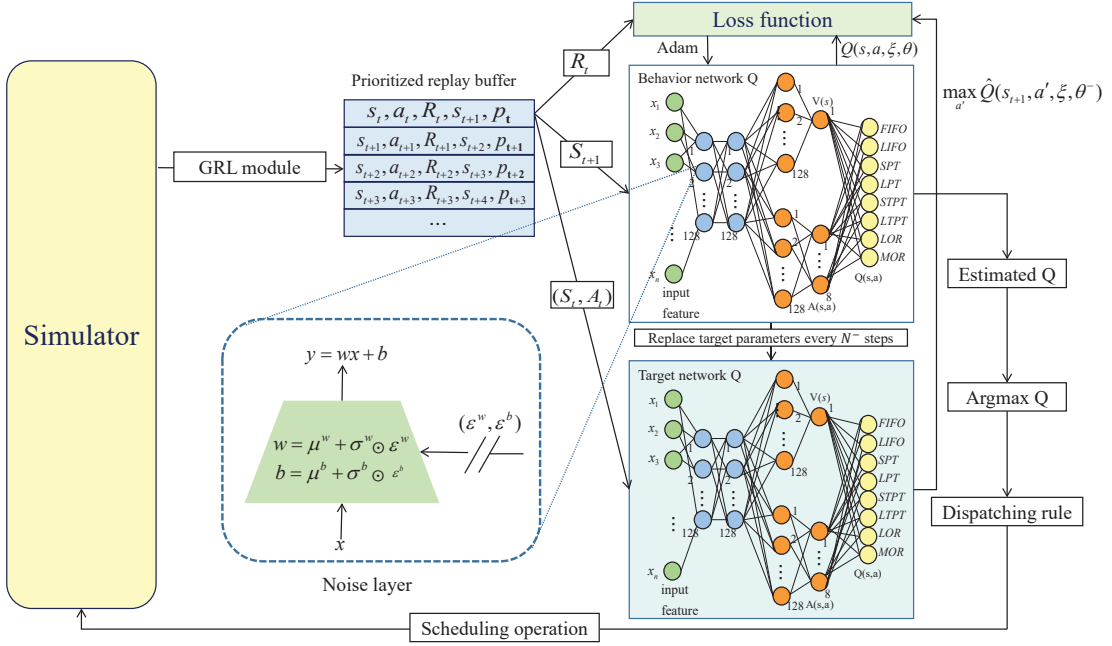}
    \caption{The double dueling DQN with prioritized replay and noisy networks.}
    \label{Figure 6}
    
\end{figure}

The architecture of D3QPN is shown in Figure \ref{Figure 6}, which integrates all the components mentioned above into a single agent.

\section{Experiments}
\label{5}
\subsection{Gymjsp}
\label{5.1}
\begin{figure}[h!]
    \centering
    \includegraphics[width=14.5cm]{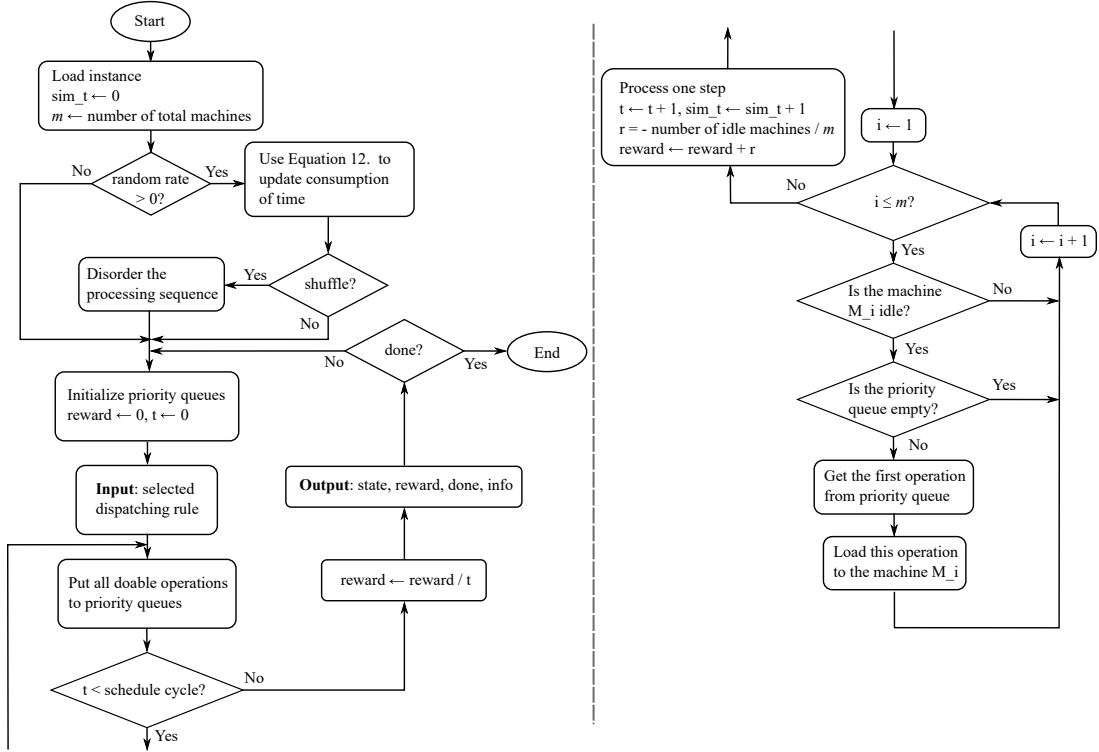}
    \caption{The scheduling process based on Gymjsp. $\textbf{sim\_t}$: the simulated global time; $\textbf{t}$: the accumulated execution time of the current rule; $\textbf{done}$: the indicator of whether all jobs have finished; $\textbf{info}$: extra information such as makespan (zero if there are still unfinished operations).}
    \label{Figure 7}
\end{figure}

As mentioned in Section \ref{2.3}, there is a critical lack of standardized practice in current RL-based JSP research in terms of experimental simulators and evaluation procedures, making it difficult for algorithm comparison. Furthermore, existing simulators are not friendly to researchers to implement their RL algorithms on JSP, which may seriously hinder the development of RL-based JSP solutions. To bridge this gap, we propose a new public benchmark named \textbf{Gymjssp}\footnote{https://github.com/Yunhui1998/Gymjsp} based on the well-known OR-Library, a collection of datasets for a variety of Operations Research (OR) problems including JSP, portfolio optimization and so on. To provide an easy-to-use suite for RL algorithms, Gymjsp features an interface similar to OpenAI Gym (Brockman et al. \citeyear{z45}), which is a widely used benchmark in the RL community. According to Figure \ref{Figure 7}, the scheduling environment is initialized based on the instances of the OR-Library and, if required, with disturbance on the time consumption (a longer time consumption than normal one of an operation on a machine can be viewed as the event of machine breakdown for the next operation). It is also possible to control whether to disorder the processing sequence of operations by the parameter $\textbf{shuffle}$, representing different order requirements. All the dynamic effects can be reproduced using the same random seeds by the function $\textbf{seed}$. During initialization, with the parameter $\textbf{random rate}$, the time consumptions of all operations are determined as follows:

\begin{gather}
T'_{i,l} = T_{i,l}+noisy_{(i,l)}\\
noisy_{(i,l)} =
\left\{  
    \begin{array}{lr}  
        \text{min}(1,\text{max}(-1,\mathcal N(0,0.1)))\cdot T_{i,l}, & r < random\_rate\\
        0, & other\\
    \end{array}  
\right.
\end{gather}
where $T'_{i,j}$ is the new time consumption for the $j^{th}$ operation of the job $J_{i}$; $r\in [0,1]$ is a random float; $random\_rate \in [0,1]$ is a hyperparameter.

After environment initialization, in each decision step, a dispatching rule from $\{\text{FIFO, LIFO, LPT, SPT, STPT, LPTP, MOR, LOR}\}$ is selected as the action. Each dispatching rule assigns a specific priority queue for each machine and, for machine $M_i$, its priority queue $q_i$ contains, in ascending order, all doable operations to be processed by $M_i$. Next, the environment schedules operations following this rule for $\textbf{schedule cycle}$ times and returns the average ratio of idle machines as the reward of this action. Finally, when all the jobs are finished, the \textbf{done} flag is set to \textbf{True}, and the extra information such as makespan is stored in \textbf{info} for debugging or testing.

\subsection{Performance evaluation}
\label{5.2}
The key objective of the experimental studies is to investigate the advantage of our proposed framework of hybrid intelligence compared with traditional dispatching rules. In particular, we show that dynamically selecting appropriate dispatching rules based on the latest problem states can provide consistent superiority over any single fixed dispatching rule. We also compare the proposed method with genetic algorithms (GAs) to demonstrate its performance from a different perspective. Furthermore, thanks to the unified interface provided by Gymjsp, it is more convenient than ever to conduct systematic studies of various RL algorithms on the DJSP. The competitive algorithms adopted in our studies are listed as follows (the same problem formulation, GRL methods, and the number of training episodes are used for all RL methods): 

\begin{itemize}
\item [1.] Eight dispatching rules: FIFO, LIFO, MOR, LOR, LPT, SPT, LTPT, STPT
\item [2.] GA (Chen et al. \citeyear{z46}): GA is a meta-heuristic technique reflecting the principles of biological evolution where the fittest individuals are selected to produce the offspring to form the new generation.
\item [3.] A2C (Chen, Hu, and Min \citeyear{z48}): Advantage actor-critic is an asynchronous policy-based RL method, using the actor to interact with the environment and the critic to criticize the actions made by the actor.
\item [4.] PPO (Schulman et al. \citeyear{z47}): Proximal policy optimization is a highly effective policy gradient RL method built upon trust region policy optimization (TRPO), which can obtain monotonic improvement by controlling the divergence between the old policy and the new policy.
\item [5.] DQN (Luo \citeyear{z9}): Deep Q-Network combines Q-learning with convolutional neural networks and experience replay, well regarded as the foundation of deep RL research.
\item [6.] Rainbow DQN (Hessel et al. \citeyear{z49}): Rainbow combines several independent extensions to DQN to provide state-of-the-art performance.
\end{itemize}

For the sake of generalization and fairness, we keep the efforts on hyperparameter tuning to the minimum. Table \ref{Table.2} summarizes the hyperparameters of D3QPN, whose values are well known to be appropriate on various tasks and are kept unchanged across all instances.

\begin{table}[]
\caption{Hyperparameters}
\label{Table.2}
\begin{tabular}{llll}
\hline
\textbf{Hyperparameters}    & \textbf{Values} & \textbf{Hyperparameters}       & \textbf{Values} \\ \hline
Number of training episodes & 3000            & Discount factor $\gamma$     & 0.99            \\
Number of testing episodes  & 500             & Prioritized replay $\alpha$       & 0.6             \\
Schedule cycle            & 8               & Prioritized replay $\beta$        & 0.4             \\
Buffer size                 & 100000          & Number of layers of GRL module & 3                \\
Step size               & 0.0003          &  Number of attention heads      & 5               \\
Batch size                  & 64               & Random rate of the environment & 0.1             \\
Target Q update frequency   & 100   & Shuffle & True                       \\ \hline
\end{tabular}
\end{table}

We measure the performance of D3QPN on 12 instances of different sizes to explore the flexibility of D3QPN. All methods are evaluated for 500 episodes, and the full results are shown in Table \ref{Table.3}. In general, our method achieves the minimum makespan on all instances except ft06. The reason is that, with $\textbf{schedule cycle}$ set to 8, our method can only make 7 or 8 times decisions on ft06, which dramatically limits its performance. If we allow D3QPN to make a decision in every step, it can achieve the same performance as the GA on ft06. Compared with dispatching rules, our proposed method achieves smaller makespan than any individual dispatching rule, with an average performance improvement of 17.80$\%$, confirming the value and superiority of hybrid intelligence over single-agent decision making. Our method also achieves smaller makespan than the meta-heuristic algorithm GA, and the average performance improvement is 28.51$\%$. Furthermore, D3QPN is more effective than other RL algorithms such as A2C, PPO, and DQN, with an average performance improvement of 11.40$\%$. An interesting finding is that D3QPN can outperform Rainbow, which contains more extensions to DQN than D3QPN. In the ablation study in Section \ref{5.5}, we will conduct further analysis to show that not all the components in Rainbow are beneficial for the DJSP.

\begin{table}[h]
  \renewcommand\arraystretch{1.3}
  \caption{The comparison of various algorithms on DJSP instances. The size of the instance is given by the number of jobs $\times$ the number of machines, and the best makespan results are printed in boldface.}
  \label{Table.3}
\resizebox{\textwidth}{!}{
\begin{tabular}{llllllllllllllll}
\hline
         &       & \multicolumn{8}{c}{Dispatching rule}                  &      & \multicolumn{5}{c}{Reinforcement Learning}   \\ \cline{3-10} \cline{12-16} 
Instance & Size  & FIFO & LIFO & LPT  & SPT  & LTPT & STPT & MOR  & LOR  & GA   & A2C  & PPO  & DQN  & Rainbow & \textbf{Ours} \\ \hline
ft06     & 6$\times$6   & 64   & 70   & 74   & 85   & 66   & 78   & 59   & 67   & \textbf{58}   & 69   & 67   & 65   & 63      & 60            \\
la01     & 10$\times$5  & 826  & 792  & 818  & 762  & 827  & 933  & 780  & 939  & 738  & 830  & 828  & 785  & 935     & \textbf{695}  \\
la06     & 15$\times$5  & 996  & 1234 & 1119 & 1181 & 1074 & 1094 & 927  & 1097 & 982  & 1043 & 1021 & 984  & 1066    & \textbf{921}  \\
la11     & 20$\times$5  & 1247 & 1496 & 1468 & 1432 & 1418 & 1493 & 1227 & 1570 & 1330 & 1225 & 1331 & 1283 & 1480    & \textbf{1202} \\
la21     & 15$\times$10 & 1304 & 1379 & 1406 & 1304 & 1338 & 1473 & 1267 & 1480 & 1502 & 1334 & 1345 & 1347 & 1494    & \textbf{1210} \\
la31     & 30$\times$10 & 1884 & 2197 & 2248 & 1981 & 2089 & 2297 & 1833 & 2196 & 2436 & 2075 & 2047 & 1958 & 1846    & \textbf{1775} \\
la36     & 15$\times$15 & 1620 & 1833 & 1799 & 1809 & 1747 & 1894 & 1490 & 1932 & 1873 & 1692 & 1682 & 1642 & 1828    & \textbf{1481} \\
orb01    & 10$\times$10 & 1374 & 1355 & 1396 & 1389 & 1299 & 1427 & 1314 & 1385 & 1432 & 1344 & 1343 & 1327 & 1473    & \textbf{1141} \\
swv01    & 20$\times$10 & 2061 & 1832 & 2202 & 1683 & 1962 & 1843 & 2008 & 1877 & 2319 & 1979 & 1986 & 1962 & 2061    & \textbf{1668} \\
swv06    & 20$\times$15 & 2455 & 2250 & 2597 & 2232 & 2312 & 2419 & 2321 & 2355 & 2960 & 2369 & 2354 & 2311 & 2333    & \textbf{2068} \\
swv11    & 50$\times$10 & 4448 & 3771 & 4624 & 3626 & 3870 & 3825 & 4602 & 3957 & 5354 & 3819 & 4103 & 3998 & 3937    & \textbf{3534} \\
yn1      & 20$\times$20 & 1158 & 1181 & 1139 & 1121 & 1152 & 1196 & 1053 & 1238 & 1496 & 1250 & 1132 & 1109 & 1110    & \textbf{1032} \\ \hline
\end{tabular}}
\end{table}

\subsection{The effect of state representation}
\label{5.3}
\begin{figure}[htp]
    \centering
    \includegraphics[width=15cm]{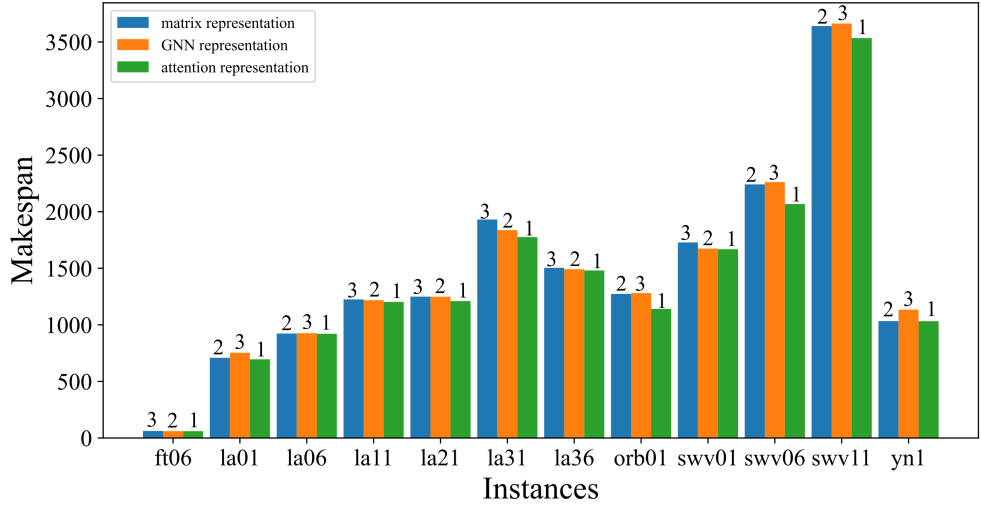}
    \caption{The Makespan performance of three GRL methods over different instances. The numbers above the bar show the ranks on each instance.}
    \label{Figure 8}
\end{figure}

\begin{figure}[htp]
    \centering
    \includegraphics[width=15cm]{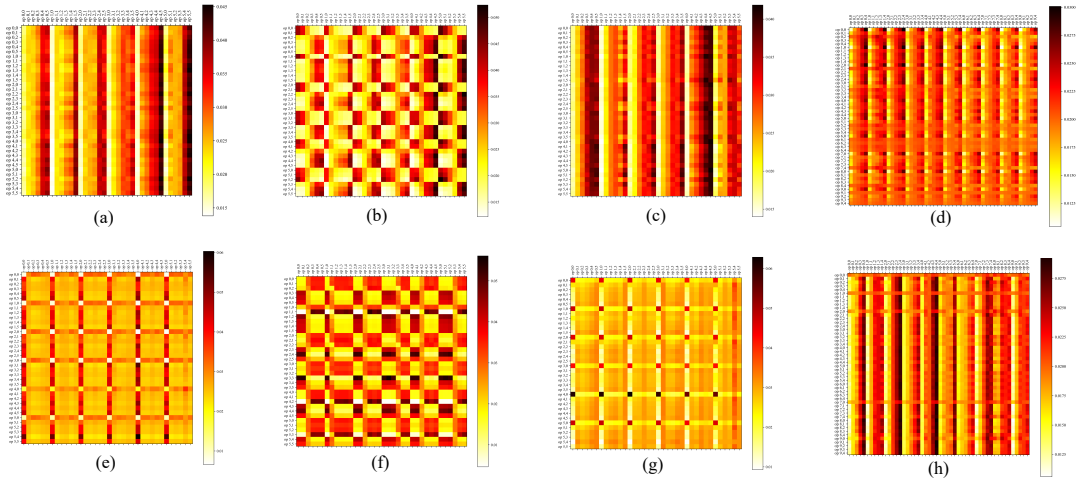}
    \caption{Visualization of attention weights at the convergence of the GRL module. (a) the attention map of the first layer when the input is the initial state of ft06; (b) the attention map of the first layer when the input is the state of ft06 after 20 steps; (c) the attention map of the first layer when the input is the initial state of ft06 with random initialization; (d) the attention map of the first layer when the input is the initial state of la01; (e) the attention map of the third layer when the input is the initial state of ft06; (f) the attention map of the third layer when the input is the state of ft06 after 20 steps; (g) the attention map of another head of the third layer when the input is the initial state of ft06. (h) the attention map of the first layer of the GRL module trained on ft06 when the input is the initial state of la01.}
    \label{Figure 9}
\end{figure}

To verify the rationality of using attention mechanism as the GRL method, we compare attention mechanism with GNN and matrix representation, and the ranking results are shown in Figure \ref{Figure 8}. It is clear that the attention mechanism achieves better performance than the GNN and the matrix representation on all instances. To gain a deep insight into its performance, we visualize the attention weights in different dimensions, and the resulting attention maps are shown in Figure \ref{Figure 9}.

Each map contains the attention weights of each operation (x-axis) relative to all other operations (y-axis). For example, the element in the first row and the second column represents the attention weight that $O_{00}$ assigns to $O_{01}$, which is the amount of attention that $O_{00}$ should pay to $O_{01}$ (the darker the square, the greater the weight). According to the visualization results, there is a list of important findings: 

\begin{itemize}
\item From (a), (c), and (d): the first layer (shallow layer) of the attention model pays more attention to the subsequent operations of each job.
\item By comparing (a) and (b): in the first layer, the unfinished operations pay more attention to the subsequent unfinished operations and ignore the completed operations; completed operations pay more attention to completed operations, which is a simple rule to determine the current processing operation of each job. 
\item By comparing (a) and (e) or (a) and (g) or (b) and (f): unlike the attention in shallow layers where each operation only focuses on the subsequent operations, deeper layers develop a more complex model that can be seen as a kind of more abstract knowledge according to the current operating state.
\item From (e) and (g): different heads of attention use different strategies to assign weights, confirming that multi-head attention is beneficial for acquiring valuable information from different angles.
\item By comparing (a), (c), and (h): it shows that the attention mechanism is robust and can adapt to dynamic events.
\end{itemize}

\subsection{The effect of action space}
\label{5.4}
\begin{figure}[htp]
    \centering
    \includegraphics[width=15cm]{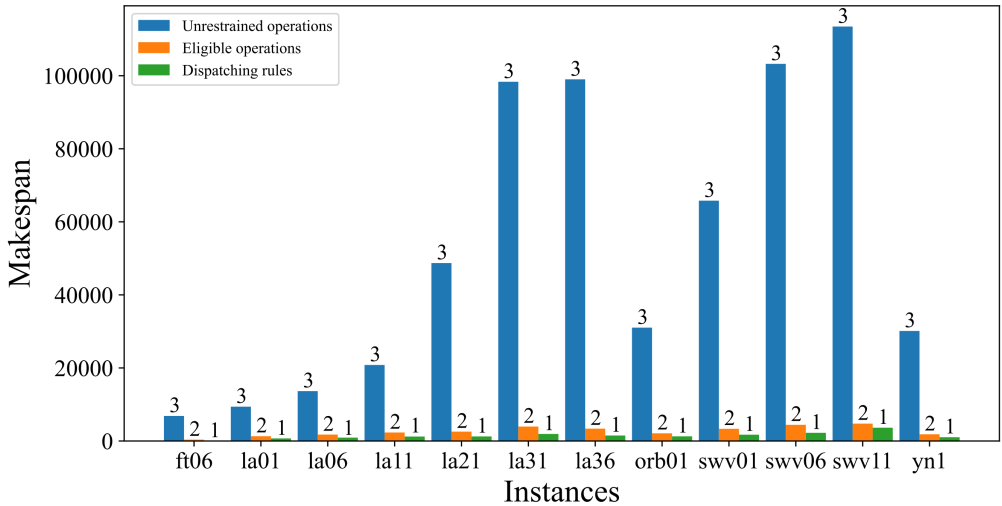}
    \caption{The Makespan performance using three types of action spaces over different instances. The numbers above the bar show the ranks on each instance.}
    \label{Figure 10}
\end{figure}

In some RL-based DJSP methods, customized dispatching rules are used as the action space due to their clarity, ease of implementation, and good interpretability. However, there is a lack of empirical evidence on the advantage over using unrestrained operations or eligible operations as the action space. Figure \ref{Figure 10} presents the experimental results on the effect of different action spaces. 

It is clear that, on the 12 instances, using dispatching rules as the actions is always better than using eligible or unrestrained operations. Moreover, using unrestrained operations as actions produce significantly worse results than others. This is because, without any restrictions on the selection of operations, the agent is likely to select a non-executable operation that has been completed, is being processed, requires an occupied machine, or whose antecedent operation has not been completed. Compared with selecting an eligible operation in each step, selecting a dispatching rule to be executed for some time cycles can ensure the consistence of operations and obtain more stable performance. More importantly, using dispatching rules as the action space provides a principled approach to unifying structured domain knowledge and data-driven intelligence with improved flexibility and efficiency.

\begin{table*}[h]
\renewcommand\arraystretch{1.3}
  \caption{The makespan performance of DQN, D3QPN and DQN variants each with single component from Rainbow. ↑ and ↓ indicate that this method has larger and smaller makespan than DQN, respectively. The results printed in boldface indicate better performance than DQN.}
  \label{Table.4}
\resizebox{\textwidth}{!}{%
\begin{tabular}{|c|c|cc|cc|cc|cc|cc|cc|cc|}
\hline
\multirow{2}{*}{Instance} & \multicolumn{15}{c|}{Makespan-Improvement} \\ \cline{2-16} 
     & DQN  & \multicolumn{2}{c|}{Double}    & \multicolumn{2}{c|}{Dueling}   & \multicolumn{2}{c|}{Per}       & \multicolumn{2}{c|}{Noisy}     & \multicolumn{2}{c|}{Cat}      & \multicolumn{2}{c|}{N\_step}  & \multicolumn{2}{c|}{\textbf{D3QPN}} \\ \hline
ft06 & 65   & \textbf{64}   & \textbf{↓1\%}  & \textbf{61}   & \textbf{↓6\%}  & \textbf{64}   & \textbf{↓2\%}  & \textbf{59}   & \textbf{↓8\%}  & \textbf{64}   & \textbf{↓1\%} & \textbf{62}   & \textbf{↓4\%} & \textbf{60}      & \textbf{↓7\%}    \\
la01 & 785  & \textbf{709}  & \textbf{↓10\%} & \textbf{780}  & \textbf{↓1\%}  & \textbf{783}  & \textbf{0\%}     & \textbf{718}  & \textbf{↓9\%}  & 844 & ↑8\% & 814 & ↑4\% & \textbf{695}     & \textbf{↓11\%}   \\
la06 & 984  & \textbf{971}  & \textbf{↓1\%}  & \textbf{965}  & \textbf{↓2\%}  & 996 & ↑1\% & \textbf{959}  & \textbf{↓3\%}  & 1076 & ↑9\% & \textbf{969}  & \textbf{↓1\%} & \textbf{921}     & \textbf{↓6\%}    \\
la11 & 1283 & \textbf{1278} & \textbf{0\%}     & \textbf{1234} & \textbf{↓4\%}  & \textbf{1238} & \textbf{↓4\%}  & \textbf{1229} & \textbf{↓4\%}  & 1299 & ↑1\% & \textbf{1240} & \textbf{↓3\%} & \textbf{1202}    & \textbf{↓6\%}    \\
la21 & 1347 & \textbf{1279} & \textbf{↓5\%}  & \textbf{1297} & \textbf{↓4\%}  & \textbf{1306} & \textbf{↓3\%}  & \textbf{1299} & \textbf{↓4\%}  & \textbf{1339} & \textbf{↓1\%} & 1351 & 0\% & \textbf{1210}    & \textbf{↓10\%}   \\
la31 & 1958 & \textbf{1928} & \textbf{↓2\%}  & \textbf{1893} & \textbf{↓3\%}  & \textbf{1914} & \textbf{↓2\%}  & \textbf{1920} & \textbf{↓2\%}  & 2092 & ↑7\% & 2110 & ↑8\% & \textbf{1775}    & \textbf{↓9\%}    \\
la36 & 1642 & \textbf{1639} & \textbf{0\%}     & 1648 & 0\% & \textbf{1570} & \textbf{↓4\%}  & \textbf{1628} & \textbf{↓1\%}  & 1705 & ↑4\% & 1885 & ↑15\% & \textbf{1481}    & \textbf{↓10\%}   \\
orb01 & 1327 & \textbf{1267} & \textbf{↓5\%}  & \textbf{1230} & \textbf{↓7\%}  & \textbf{1195} & \textbf{↓10\%} & \textbf{1161} & \textbf{↓12\%} & 1374 & ↑4\% & 1355 & ↑2\% & \textbf{1141}    & \textbf{↓14\%}   \\
swv01 & 1962 & \textbf{1772} & \textbf{↓10\%} & \textbf{1770} & \textbf{↓10\%} & \textbf{1770} & \textbf{↓10\%} & \textbf{1711} & \textbf{↓13\%} & 2080 & ↑6\% & \textbf{1933} & \textbf{↓2\%} & \textbf{1668}    & \textbf{↓15\%}   \\
swv06 & 2311 & \textbf{2244} & \textbf{↓3\%}  & \textbf{2144} & \textbf{↓7\%}  & 2337 & ↑1\% & \textbf{2299} & \textbf{↓1\%}  & 2339 & ↑1\% & 2377 & ↑3\% & \textbf{2068}    & \textbf{↓11\%}   \\
swv11 & 3998 & \textbf{3881} & \textbf{↓3\%}  & \textbf{3632} & \textbf{↓9\%}  & \textbf{3635} & \textbf{↓9\%}  & \textbf{3697} & \textbf{↓8\%}  & \textbf{3891} & \textbf{↓3\%} & \textbf{3661} & \textbf{↓8\%} & \textbf{3534}    & \textbf{↓12\%}   \\
yn1  & 1109 & \textbf{1089} & \textbf{↓2\%}  & \textbf{1047} & \textbf{↓6\%}  & \textbf{1055} & \textbf{↓5\%}  & 1124 & ↑1\% & \textbf{1095} & \textbf{↓1\%} & 1114 & 0\% & \textbf{1032}    & \textbf{↓7\%}    \\ \hline
Average & 1564 & \textbf{1510} & \textbf{↓3\%}  & \textbf{1475} & \textbf{↓5\%}  & \textbf{1488} & \textbf{↓4\%}  & \textbf{1484} & \textbf{↓5\%}  & 1600 & ↑3\% & 1573 & ↑1\% & \textbf{1399}    & \textbf{↓10\%}   \\ \hline
\end{tabular}%
}
\end{table*}

\subsection{Ablation study}
\label{5.5}
As shown in Section \ref{5.2}, the performance of D3QPN is significantly better than DQN and Rainbow. To better understand the contribution of each component in D3QPN, we perform an ablation study in which we investigate the performance of a number of DQN variants each with a single component in Rainbow. The experiment results are shown in Table \ref{Table.4}.

We can see that not all the variants in Rainbow achieve better performance than DQN, which means that some of them are not suitable for the DJSP. For example, distributional DQN (Bellemare, Dabney, and Munos \citeyear{z54}) and multi-step DQN (Peng and Williams \citeyear{z55}) achieve higher makespan values than DQN in general, although multi-step DQN performs better than DQN on some instances. This explains why Rainbow's performance is worse than that of D3QPN: the multi-step learning and distributional network are negative factors on Rainbow. We hypothesize that this is because the distributional network inherently provides the agent with a distributional perspective on action values, but there are just a few actions to choose from in the DJSP, creating more interference than necessary. Furthermore, the multi-step learning considers the reward before many steps. However, we would prefer the agent to focus on the immediate reward, the machine utilization, rather than the previous reward. In contrast, each component in D3QPN improves the performance of DQN, with NoisyDQN and DuelingDQN, in particular, reducing makespan by 5$\%$. 

\begin{figure}[h!]
    \centering
    \includegraphics[width=14cm]{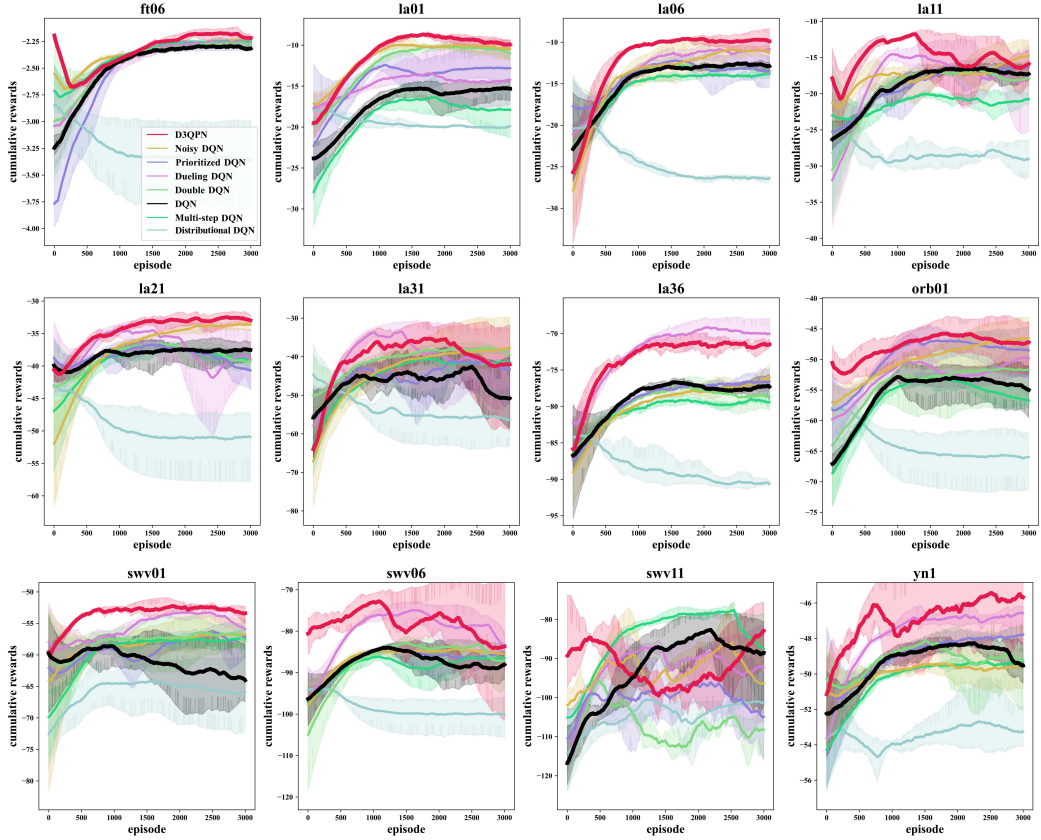}
    \caption{The training curves of DQN, D3QPN, and DQN variants each with a single component from Rainbow. The vertical axis of the graph shows the method's cumulative reward every episode, reflecting the performance of each method. The solid lines and shaded regions represent the mean and standard deviation, respectively, across five runs.}
    \label{Figure 11}
 
\end{figure}

We also visualize the cumulative rewards during the training of each method to further support our conclusion. The results in Figure \ref{Figure 11} show that D3QPN achieves maximal cumulative rewards on almost all instances with a stable decision-making policy. Also, the components in D3QPN help DQN achieve better performance, while the performance of the distributional DQN becomes worse as the training goes on.

\section{Conclusion}
\label{6}
In this paper, we propose a flexible RL-based framework for solving DJSP tasks, which takes disjunctive graphs as states and a set of general dispatching rules as the action space. It is a promising paradigm of hybrid intelligence as it combines structured human knowledge in the form of dispatching rules and flexible data-driven machine intelligence. Moreover, we use the attention mechanism as the graph representation learning module for effective feature extraction and present D3QPN, an improved DQN algorithm, to map each DJSP state to the most appropriate dispatching rule. To promote the application of RL in the JSP, we present a public benchmark Gymjsp based on the well-known OR-Library, which provides a standardized off-the-shelf and easy-to-use suite for RL communities. Experimental results based on Gymjsp show that our proposed method can outperform traditional rule-based decision systems, meta-heuristic techniques, and SOTA RL algorithms, with great potential in the field of smart manufacturing.

As to future work, it is worthwhile to extend our framework to other variants of JSP, such as the FJSP. Due to the flexibility of our framework, it is also possible to investigate various novel JSP scenarios where the machines or jobs may exhibit unprecedented levels of dynamics. Furthermore, specialized domain knowledge can be strategically integrated into our framework to, for example, design more indicative reward functions or extract more valuable state features.

\section*{Disclosure statement}
No potential conflict of interest was reported by the author(s).

\section*{Data Availability Statement}
The data that support the findings of this study are openly available in https://github.com/Yunhui1998/Gymjsp.

\section*{Funding}
This work is supported by the National Natural Science Foundation of China (U1713214).

\section*{Notes on contributor(s)}
\textbf{Yunhui Zeng}: Methodology, Conceptualization, Writing - Original Draft, Coding, Formal analysis. \textbf{Zijun Liao}: Coding, Writing - Review \& Editing, Formal analysis. \textbf{Yuanzhi Dai}: Investigation, Writing - Review \& Editing. \textbf{Rong Wang}: Investigation, Writing - Review \& Editing. \textbf{Xiu Li}: Supervision, Writing - Review \& Editing. \textbf{Bo Yuan}. Resources, Supervision, Writing - Review \& Editing.

\bibliography{ref}
\bibliographystyle{tfcad}
\end{document}